\documentclass[letterpaper, 10 pt, conference]{ieeeconf}  

\IEEEoverridecommandlockouts                              

\usepackage{graphicx}
\usepackage{color,soul}
\usepackage{dblfloatfix}
\usepackage{multirow}
\usepackage{multicol}
\usepackage{hyperref}
\usepackage{siunitx}
\usepackage{hyperref}
\usepackage{xcolor}
\usepackage{graphicx}
\usepackage{svg}
\usepackage{amsmath}
\usepackage{amsfonts}
\usepackage{amssymb}
\usepackage{array}
\usepackage{url}
\usepackage{epstopdf}
\usepackage{float}
\usepackage{subfigure}
\usepackage{threeparttable}
\usepackage{bm}
\usepackage{caption}
\usepackage{algorithm,algorithmic}
\usepackage{booktabs}
\usepackage{tabularx}
\usepackage[utf8]{inputenc} 
\usepackage[T1]{fontenc}    
\usepackage{hyperref}       
\usepackage{url}            
\usepackage{booktabs}       
\usepackage{amsfonts}       
\usepackage{nicefrac}       
\usepackage{microtype}      
\usepackage{xcolor}         
\usepackage{graphicx}
\usepackage{multirow}
\usepackage{graphicx}
\usepackage{bbding}
\usepackage{colortbl}
\usepackage{CJKutf8}
\usepackage{cite}
\usepackage{subcaption} 

\begin{document}

\definecolor{darkgreen}{rgb}{0.0, 0.5, 0.0}
\definecolor{darkred}{rgb}{0.5, 0.0, 0.0}

\newcommand{\etal}{\textit{et al}.}


\title{\LARGE \bf AssistantX: An LLM-Powered Proactive Assistant in Collaborative Human-Populated Environments
}


\author{Nan Sun$^{1,*}$, Bo Mao$^{2,*}$,  
 Yongchang Li$^{1,*}$, Di Guo$^{2}$ and Huaping Liu$^{1,3}$%
\thanks{$^{1}$The author is with the Department of Computer Science and Technology, Tsinghua University, Beijing, 100084, China.}%
\thanks{$^{2}$The author is with the School of Artificial Intelligence,  Beijing University of Posts and Telecommunications,Beijing, 100876, China.}%
\thanks{$^{3}$Corresponding Author. hpliu@tsinghua.edu.cn}%
\thanks{$^{*}$Equal Contribution.}%
}

\maketitle
\thispagestyle{empty}
\pagestyle{empty}


\begin{abstract}
Current service robots suffer from limited natural language communication abilities, heavy reliance on predefined commands, ongoing human intervention, and, most notably, a lack of proactive collaboration awareness in human-populated environments. This results in narrow applicability and low utility. In this paper, we introduce AssistantX, an LLM-powered proactive assistant designed for autonomous operation in real-world scenarios with high accuracy. AssistantX employs a multi-agent framework consisting of 4 specialized LLM agents, each dedicated to perception, planning, decision-making, and reflective review, facilitating advanced inference capabilities and comprehensive collaboration awareness, much like a human assistant by your side. We built a dataset of 210 real-world tasks to validate AssistantX, which includes instruction content and status information on whether relevant personnel are available. Extensive experiments were conducted in both text-based simulations and a real office environment over the course of a month and a half. Our experiments demonstrate the effectiveness of the proposed framework, showing that AssistantX can \textit{reactively} respond to user instructions, \textit{actively} adjust strategies to adapt to contingencies, and \textit{proactively} seek assistance from humans to ensure successful task completion. More details and videos can be found at \url{https://assistantx-agent.github.io/AssistantX/}.
\end{abstract}

\section{INTRODUCTION}

Imagine having a capable assistant; you would naturally expect it to handle various tasks on your behalf. For example, if you need to print a file but lack access to a printer, your expectation is simply to send the file to the assistant, which will
manage the rest—locating someone with or near a printer, requesting them to print it, and ultimately returning the printed document to you. You would only need to receive the file, with any uncertainties handled autonomously by the assistant. This highlights the need for an agentic system that can respond to diverse user needs, aligning both physical and digital environments to enhance efficiency \cite{liu2024aligningcyberspacephysical}.

\begin{figure}[t]
    \centering
    \includegraphics[width=1\linewidth]{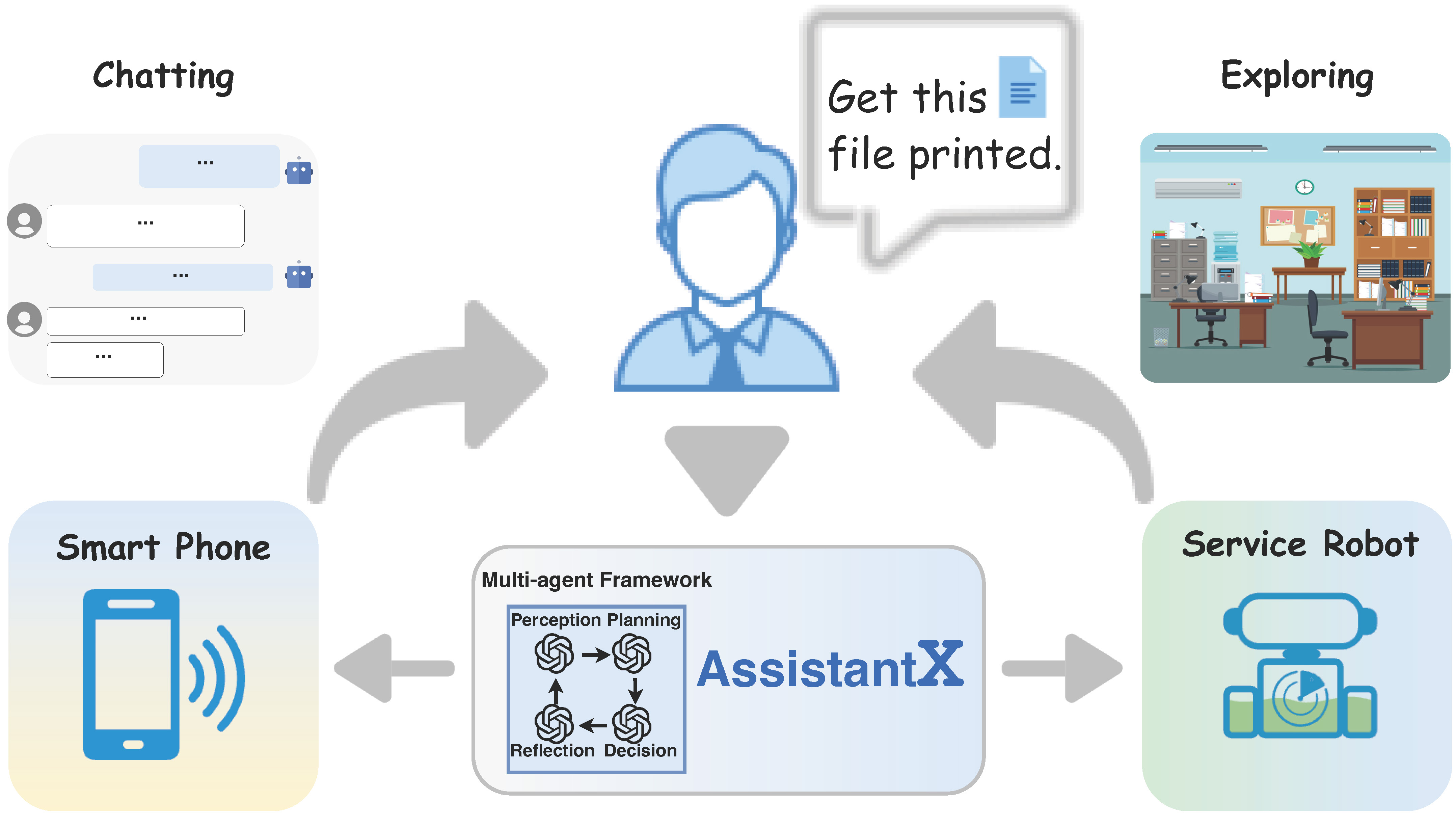}
    \caption{AssistantX overcomes the limitations of existing service robots and virtual assistants, seamlessly integrating physical and virtual actions to meet human needs.}
    \vspace{-7mm}
    \label{fig:introduction1}
\end{figure}

\begin{figure*}[h]
    \centering
    \includegraphics[width=1\linewidth]{image617/problem_for617.jpg}
    \caption{AssistantX proficiently generates both cyber tasks $\mathcal{TC}$ and real-world tasks $\mathcal{TR}$, executing them concurrently in a manner akin to a human assistant. This approach enhances efficiency and promotes productivity.}
    \vspace{-7mm}
    \label{fig:problem_formulation}
\end{figure*}

However, service robots we encounter in daily life often rely on predefined commands, requiring direct physical interaction, and lack proactive engagement. A recent development, the LLM-integrated Assistant Robot — OfficeMate, demonstrates a few breakthroughs in human-robot interaction in office settings \cite{pan2025officematepilotevaluationoffice}, but it still depends on direct voice commands or keyboard inputs. Despite integrating GPT-4 for speech-to-text conversion, it only maps user input to predefined workflows using a trigger word database, lacking truly proactive behaviors. Issues like ineffective long-distance communication in real-time, inability to handle unexpected situations, and lack of collaboration with humans on complex tasks remain unsolved, limiting its potential as a fully functional assistant.

These limitations motivate the development of \textbf{AssistantX}: an LLM-powered proactive assistant with a virtual presence in cyberspace and a physical embodiment in the real world, designed to interact more effectively with humans for task execution (see Fig. \ref{fig:introduction1}). Built on a multi-agent framework-\textbf{PPDR4X}, AssistantX features four specialized LLM agents capable of \textbf{P}erceiving, \textbf{P}lanning, \textbf{D}eciding actions, and \textbf{R}eflecting on outcomes. Seamlessly integrated into daily workflows, AssistantX functions like a human assistant by your side, bridging the gap between cyber and physical interactions (see Fig. \ref{fig:problem_formulation}). We demonstrate through extensive experiments that AssistantX can \textit{reactively} respond to user instructions, \textit{actively} adapt its strategies to contingencies, and \textit{proactively} seek human assistance to ensure task success.

Our contributions include:
\begin{enumerate}
  \item We propose the LLM-powered multi-agent framework, \textbf{PPDR4X} (Perception, Planning, Decision, Reflection for AssistantX), which significantly enhances robotic cognition and elevates problem-solving capabilities.
  
  \item We develop \textbf{AssistantX}, a proactive assistant that combines the strengths of digital and embodied agents to autonomously fulfill user demands by performing actions in both cyberspace and physical environments.
  
  \item We demonstrate the effectiveness of the proposed framework in human-involved tasks through extensive simulations and real-world experiments, overcoming the limitations of existing service robots.
\end{enumerate}

The structure of this paper is as follows: Section \ref{sec:realted} reviews related works, Section \ref{sec:problem} formulates the problem, Section \ref{sec:method} details the proposed framework, Section \ref{sec:experiment} presents extensive experimental results and comprehensive evaluation, and Section \ref{sec:conclusion} concludes the paper.

\begin{figure*}[t]
    \centering
    \includegraphics[width=1.0\linewidth, height=0.35\textwidth]{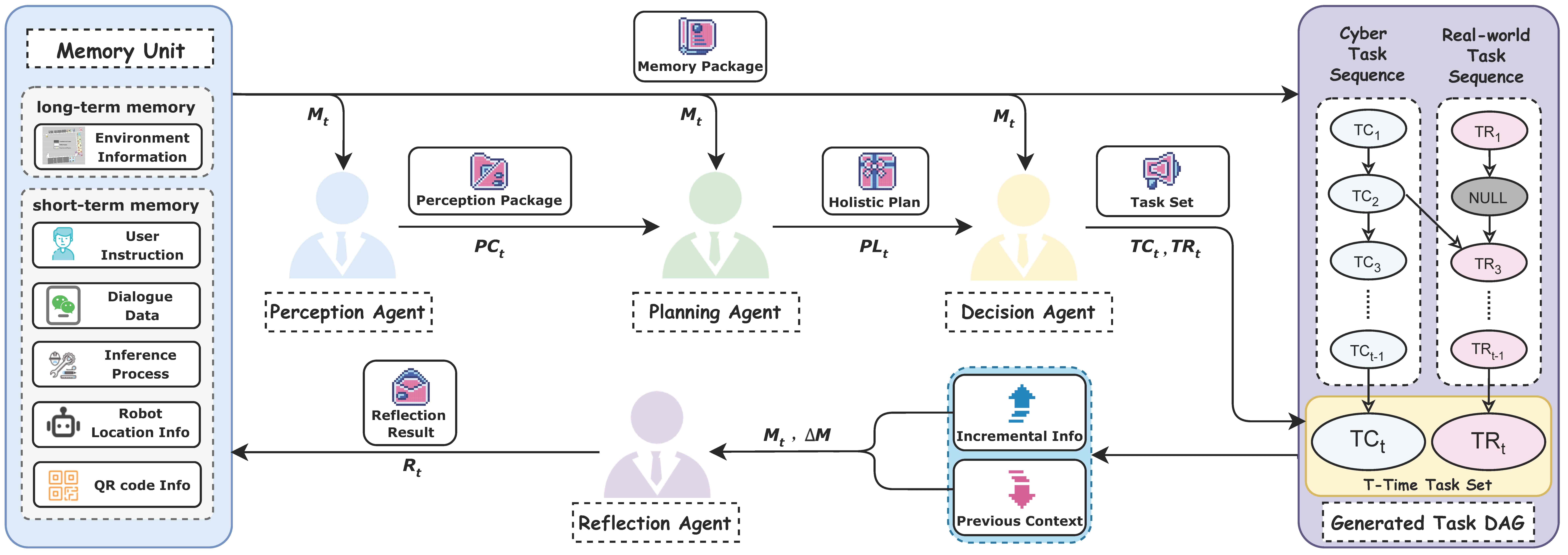}
    \caption{Overview of the proposed framework.}
    \vspace{-5mm}
    \label{fig:method}
\end{figure*} 

\section{RELATED WORK}\label{sec:realted}

\subsection{Mobile Robots in Human-Populated Environments}
Mobile robots in human-populated environments have garnered significant attention in robotics research. Early works were limited to structured settings with minimal human interaction, while recent studies emphasize adaptability and human-robot interaction in dynamic environments.

Autonomous robots that gather and report environmental data to assist humans have been developed, with an emphasis on perception rather than action \cite{chung2016}. Several studies have explored robust navigation in complex environments, but these systems struggle with real-time adaptability when unexpected situations or changes in human behavior arise \cite{zhang8570804, trautman5654369}. Methods to track human positions from dialogue have been proposed to enhance situational awareness, but their reliance on speech makes real-time decision-making challenging \cite{Liang10342509}. The recent LLM-integrated office assistant robot, OfficeMate \cite{pan2025officematepilotevaluationoffice}, shows progress in human-robot interaction but still depends on voice commands or typing on a mounted computer, limiting its usability for effective long-distance communication in real-time. Its reliance on predefined actions and trigger words restricts its ability to handle complex tasks autonomously, falling short in enhancing efficiency and bringing up productivity.

\subsection{LLM-Powered Agentic Systems}
Recent advancements in agentic systems, particularly with LLM integration, have significantly improved reasoning performance especially under Chain-of-Thought mechanism \cite{kojima2023largelanguagemodelszeroshot}. Methods like ReAct \cite{yao2023reactsynergizingreasoningacting} and  Reflexion \cite{shinn2023reflexionlanguageagentsverbal} enhance reasoning by step-by-step thinking, while multi-agent frameworks like Mobile-Agent-v2 \cite{wang2024mobileagentv2mobiledeviceoperation} and AppAgent \cite{zhang2023appagentmultimodalagentssmartphone} have been effectively applied to tasks like GUI operations on smart devices for long-horizon tasks. Additional studies further demonstrate improved human-agent interactions, decision-making, and intelligence sharing in multi-agent settings for user-centric tasks \cite{zhang2024askbeforeplanproactivelanguageagents, ren2023robotsaskhelpuncertainty, abdelnabi2024cooperationcompetitionmaliciousnessllmstakeholders, chen2024internetagentsweavingweb}. However, these approaches are primarily deployed in digital environments, with limited integration into physical embodiments, which restricts their ability to fully assist users in real-world scenarios.

Nonetheless, the results presented in these works highlight the promising potential of LLM-powered multi-agent collaboration in taking a further step toward interacting with the real world. Consequently, leveraging the synergy between digital and embodied agents to better meet human needs is becoming a key focus \cite{liu2024aligningcyberspacephysical}, which is also the core concept of our proposed AssistantX.

\section{PROBLEM FORMULATION}\label{sec:problem}

Given an office environment $\mathcal{E}$, we assume it contains $J$ distinct working locations, denoted as $\mathcal{L} = \{l_1, \dots, l_J\}$, and a set of individuals $\mathcal{H} = \{h_1, \dots, h_N\}$, where $N$ is the total number of people. The location of the $i$-th person is denoted as $Loc(h_i) \in \mathcal{L}$. We focus on the general problem of providing assistance by AssistantX to any individual in $\mathcal{H}$. Upon a request from person \( h_i \), AssistantX should navigate to public facilities (e.g., printer, fridge, coffee machine) while coordinating human assistance for completing complex tasks (e.g., printing, retrieving food, and making coffee), or deliver/retrieve items from specific individuals (e.g., delivering a file or bringing a pen) and send corresponding notifications online. We define the set of public facilities as $\mathcal{PF} = \{pf_1, \dots, pf_K\}$, where $K$ is the total number of facilities, and $Loc(pf_k) \in \mathcal{L}$ denotes the location of $pf_k$. We assume each person owns several personal items, which can be borrowed or delivered by AssistantX. The set of personal items is defined as $\mathcal{PI} = \{pi_1, \dots, pi_M\}$, where $M$ is the total number of personal items, and $Own(pi_m) = h_i \in \mathcal{H}$ indicates that person $h_i$ owns item $pi_m$. Humans in \(\mathcal{H}\) can send private messages to AssistantX to issue instructions \(\mathcal{I}\). We categorize the tasks that AssistantX can perform into two distinct types: \textit{(1) Cyber Task:} Involving operations carried out within a virtual environment by its digital avatar, such as sending notifications, making inquiries, forwarding files, and sharing QR codes, referred to as $\mathcal{TC}$; \textit{(2) Real-World Task:} Comprising tasks that require physical actions by its robotic embodiment in the real world, such as navigating to a specific location, referred to as $\mathcal{TR}$.

Once the instruction begins execution, any dialogue information between all contacts and AssistantX, including conversations occurring in any group chat involving AssistantX, will be recorded and is defined as \( \mathcal{D} \). AssistantX should perform a list of \( \mathcal{TC} \) and \( \mathcal{TR} \) based on the initial instruction \( \mathcal{I} \), the office environment \( \mathcal{E} \), and dialogue information \( \mathcal{D} \), ultimately completing the instructions. The process can be formalized as:
\begin{equation}
    (\mathcal{TC}, \mathcal{TR}) = I(\mathcal{I}, \mathcal{E}, \mathcal{D})
\end{equation}
where $I(\cdot)$ represents the LLM-powered inference process, which is the core reasoning system we aim to develop.

\section{METHODOLOGY}\label{sec:method}

In this section, we provide an overview of the framework of our proposed PPDR4X method. It comprises four specialized agents: Perception Agent, Planning Agent, Decision Agent, and Reflection Agent, each of which is built on a foundational LLM with well-crafted prompts to guide their reasoning processes. The operation of PPDR4X is iterative, with agents functioning in a loop structure, taking inputs from upstream and providing outputs to downstream (as shown in Fig. \ref{fig:method}). We also design a Memory Unit, shared by all agents, to store environmental information, online messages, and the inference process during reasoning. 

\begin{figure}[h]
    \centering
    \includegraphics[width=1\linewidth]{image617/memoryinit617.jpg}
    \caption{Overview of the data stored in Memory Unit.} 
    \vspace{-7mm} 
    \label{fig:memory unit}
\end{figure}

\begin{figure*}[h]
    \centering
    \includegraphics[width=1\linewidth]{image617/agent_thought617.jpg}
    \caption{An illustration of the inputs and outputs of PPDR agents, showing how they collaborate to determine the next move after the previous task, with all agents communicating in natural language, ensuring logical consistency and interpretability.} 
    \vspace{-6mm} 
    \label{fig:agent thought}
\end{figure*} 

\subsection{Memory Unit}

Memory Unit forms the foundation of the entire framework, storing both long-term dynamic environmental data \( \mathcal{E} \) and short-term memory  (as shown in Fig. \ref{fig:memory unit}). The short-term memory includes user instructions \( \mathcal{I} \), dialogue data \( \mathcal{D} \), embodied states \( \mathcal{S} \) (such as the robot's current location and locker state), and the inference process \( \mathcal{P} \) (which includes agent thoughts, executed tasks \( \mathcal{TC} \) and \( \mathcal{TR} \)). Long-term memory is represented using an undirected topological graph, where nodes represent humans, public facilities, personal items, and locations. The edges between nodes define relationships, such as the location of humans and facilities, and item ownership. Human nodes also include an availability attribute, indicating whether they are available to collaborate on tasks. Reflection Agent will update the edges and human node states during task execution to reflect changes in the dynamic environment. Task-specific information stored in short-term memory will also be updated during execution and reset after the completion of each user instruction. We denote the memory package at time step \( t \) as \( \mathcal{M}_t \), which is converted into descriptive text and provided as part of the input to the foundational LLM of each agent as contextual information.

\subsection{Perception of Focus Content}

If \( t \neq 0 \), given a memory package \( \mathcal{M}_t \), which contains \( \mathcal{I} \), the previous plan \( \mathcal{PL}_{t-1} \), the completed tasks \( \mathcal{TR}_{t-1} \), \( \mathcal{TC}_{t-1} \), and the reflection result \( \mathcal{R}_{t-1} \), Perception Agent is prompted to generate a fine-grained text description of the focus content (see Fig. \ref{fig:agent thought}). This includes a descriptive observation of the local environment, location and ownership information of individuals and items involved in upcoming tasks, as well as the active chat group or person in cyberspace, to better guide the next plan. The perceptual process can be articulated as follows:
\begin{equation}
    \mathcal{PC}_t = perceive(\mathcal{M}_t)
\end{equation}
where \( perceive(\cdot) \)  is the LLM-powered inference process with tailored prompts containing an output template to generate the parts of the focus content.

While we provide \( \mathcal{M}_t \) to subsequent reasoning agents as holistic contextual information, $\mathcal{PC}_t$ remains indispensable for long-horizon tasks involving a lengthy sequence of steps across multiple people and objects, which always leads to significant challenges on the planning of LLMs \cite{valmeekam2023planningabilitieslargelanguage}. In our ablation experiments, when Perception Agent is ablated, the Success Rate and Completion Rate of L3 complexity tasks decrease significantly (see TABLE \ref{tab:ablation}).
\begin{table}[h] \large
	\centering
     \caption{Action Xpace}
    \renewcommand{\arraystretch}{1.2}
    \resizebox{\columnwidth}{!}{%
    \begin{threeparttable}

    \begin{tabularx}{\textwidth}{X|X|X X X X}
      \hline
      \toprule
        Action Xpace & Action's name & \multicolumn{4}{c}{Action's Description} \\
      \hline
      
        \multirow{5}{*}{Cyber Actions}
        
        & \multicolumn{1}{l|}{\textbf{Inform} ($contact, content$)} & \multicolumn{4}{X}{Inform a contact.}
        \\
    
          \cline{2-6}
        & \multicolumn{1}{l|}{\textbf{Inquire} ($contact, question$)} & \multicolumn{4}{X}{Ask the contact a question.}\\
        
        \cline{2-6}
        & \multicolumn{1}{l|}{\textbf{Forward} ($source$, $target$)} & \multicolumn{4}{X}{Forward an electronic file from the source contact to the target contact.}\\
         \cline{2-6}
        & \multicolumn{1}{l|}{\textbf{Send QR code} ($contact$)} & \multicolumn{4}{X}{Send a QR code to the contact.}\\
        \cline{2-6}
        & \multicolumn{1}{l|}{\textbf{Wait} ($content$)} &  \multicolumn{4}{X}{Wait for someone for online message.}\\
      \hline
      
       \multirow{2}{*}{Real-World Actions}
        & \multicolumn{1}{l|}{\textbf{Move} ($proxy$ $name$)} & \multicolumn{4}{X}{Move to a location.}\\
          \cline{2-6}
        & \multicolumn{1}{l|}{\textbf{Wait in Place ($user$)}} & \multicolumn{4}{X}{Wait at the current user location for physical interaction.} \\
      \hline

       \multirow{1}{*}{Generic Actions}
        & \multicolumn{1}{l|}{\textbf{Stop}} & \multicolumn{4}{X}{Execution Terminated.}\\
    \bottomrule
     \hline
     
     \end{tabularx}
     \begin{tablenotes}    
        \normalsize          
        \item[*] We encapsulate the initial communication API provided by the social platform to send online messages, mapping digital accounts to the real names of individuals. For QR code, we first encrypt the initial QR code using the RC4 algorithm and integrate it with the smart lock system, setting a scan limit (i.e., the QR code expires after a single scan) to enhance the security of item transport. We use LiDAR to scan the floor and apply the ROS gmapping algorithm to build a semantic map. We establish a ``coordinate-name'' mapping for anchor points, where fixed locations' (x, y) coordinates are mapped to individuals' real names. Upon triggering a physical action, the robot converts the name into the corresponding coordinates and navigates to the specified location, supported by ROS Libraries' Navigation and MoveIt libraries.   
    \end{tablenotes}     
    \vspace{-9mm}
    \end{threeparttable}
     }
    \label{tab:Xpace}
 \end{table}

\subsection{Hierarchical Problem Solving}

To address long-horizon task execution in dynamic environments, we propose a hierarchical problem-solving architecture featuring an explicit separation of high-level planning and low-level action execution. Planning Agent is first prompted to creates a strategic roadmap $\mathcal{PL}_t$ based on the current memory state $\mathcal{M}_t$ and perception package $\mathcal{PC}_t$ (see Fig. \ref{fig:agent thought}):
\begin{equation}
    \mathcal{PL}_t = plan(\mathcal{M}_t, \mathcal{PC}_t)
\end{equation}
where \( plan(\cdot) \) represents the generating process of the LLM with sophisticated examples provided in prompts for in-context learning, helping with outputs that summarize completed actions to track instruction progress and provide high-level planning to guide actions.

Decision Agent then translates high-level objectives into immediate actions and operate them on the phone or dispatches the robot (see Fig. \ref{fig:agent thought}). We define an Action Xpace, comprising atomic actions for Decision Agent to select (see Table \ref{tab:Xpace} for the actions and detailed implementation of how actions operate in both cyberspace and the real world). The decision-making process is defined as:
\begin{equation}
(\mathcal {TC}_t,\! \mathcal{TR}_t) = decide(\mathcal M_t, \mathcal {PC}_t, \mathcal PL_t)
\end{equation}
where $decide(\cdot)$ denotes the mapping process utilizing the foundational LLM with dedicated prompts containing conditional constraints, detailing the dependencies of certain tasks, such as sending an online notification when moving to someone's location, or confirming that a document has been obtained or printed before asking someone to sign it. 

By incorporating higher-level reasoning to guide low-level actions, our approach ensures better alignment with the evolving needs of the task, outperforming end-to-end methods like CoT \cite{kojima2023largelanguagemodelszeroshot} and ReAct \cite{yao2023reactsynergizingreasoningacting} that directly generate actions, as their lack of a global plan limits their ability to take adaptive actions across the entire task horizon. In ablation experiments, without Planning Agent, our method suffers a degradation of over 30\% in both Cyber Task Accuracy and Real-world Task Accuracy for L2 complexity tasks and L3 (see TABLE \ref{tab:ablation}).

\subsection{Self-Reflection Mechanism} 
After the execution of $\mathcal{TC}_t$ and $\mathcal{TR}_t$, corresponding information changes or emerges, such as newly generated dialogue data from cyberspace via the cyber actions or new messages from human contacts, and the updates of robot's embodied states, including location and the smart lock signal information. To monitor these short-term data, we have set up a thread on a cloud server to track new messages through the social platform's message retrieval API, providing real-time updates. We use the ROS AMCL for robot localization, determining the robot's current position. Additionally, we have integrated a feedback circuit into the smart lock, allowing us to retrieve the magnetic lock's open/close status through a digital signal. We listen to data from the Ethernet interface to receive real-time feedback from the smart lock. The incremental information (denoted as $\Delta \mathcal{M}$) acquired will be reflected by Reflection Agent by comparing it with $\mathcal{M}_t$ and adds the insights in reflection to Memory Unit. The reflective procedure is denoted as the following formula:
\begin{equation}
    \mathcal R_t = reflect( \mathcal M_t, \mathcal {PC}_t, \mathcal {PL}_t, \mathcal {TC}_t, \mathcal {TR}_t, \Delta \mathcal{M})
\end{equation}
where $reflect(\cdot)$ represents the reflective process of the LLM, which is prompted to carefully consider whether the results of the executed action align with the expected outcomes, and then generates a reflection result $\mathcal{R}_t$ (see Fig. \ref{fig:agent thought}). It contains a binary judgment — either `Y' or `N'. A `Y' indicates Reflection Agent considers the action's outcome consistent with expectations, while `N' signals a deviation. The reflection result also summarizes the reasoning behind this judgment, providing guidance for future planning and decision-making.

This self-reflection mechanism significantly improves the Success Rate by 24\% compared to the ablated version in L3 complexity tasks (see TABLE \ref{tab:ablation}), by avoiding getting trapped by past errors. It works efficiently in tasks involving a larger number of people and facilitates proactive actions when certain personnel are unavailable.

\begin{table}[h] \small
\scriptsize
\centering
\vspace{-2mm}
\caption{Details of our dataset}
\label{tab:dataset}
    \renewcommand{\arraystretch}{0.8}
    \resizebox{\columnwidth}{!}{%
\begin{tabular}{cccc}
\toprule
    &  \multicolumn{1}{c}{\textbf{Difficulty Level}}  &\textbf{Achievability}  & \multicolumn{1}{c}{\textbf{Number}}  \\ 

\midrule
\multirow{4}{*}[0ex]{\textbf{Dataset}} & \multirow{1}{*}[0ex]{$ L_1 $}   & Achievable & 90(43\%)  \\
\cmidrule(lr){2-4}
& $ L_2 $ & \multirow{1}{*}[0ex]{Achievable}    & 73(35\%)  \\
\cmidrule(lr){2-4}
& \multirow{2}{*}[0ex]{$ L_3 $}  & Achievable   & 25(12\%)  \\
\cmidrule(lr){3-4}
&  & Unachievable   & 22(10\%)  \\
\midrule
 \textbf{Total} &  & \multicolumn{2}{c}{\textbf{210}} \\
\bottomrule
\end{tabular}
}
\vspace{-3mm}
\end{table}

\section{EXPERIMENT}\label{sec:experiment}
\subsection{Experimental Setup}
\begin{figure}[t]
    \centering
    \includegraphics[width=1.0\linewidth]{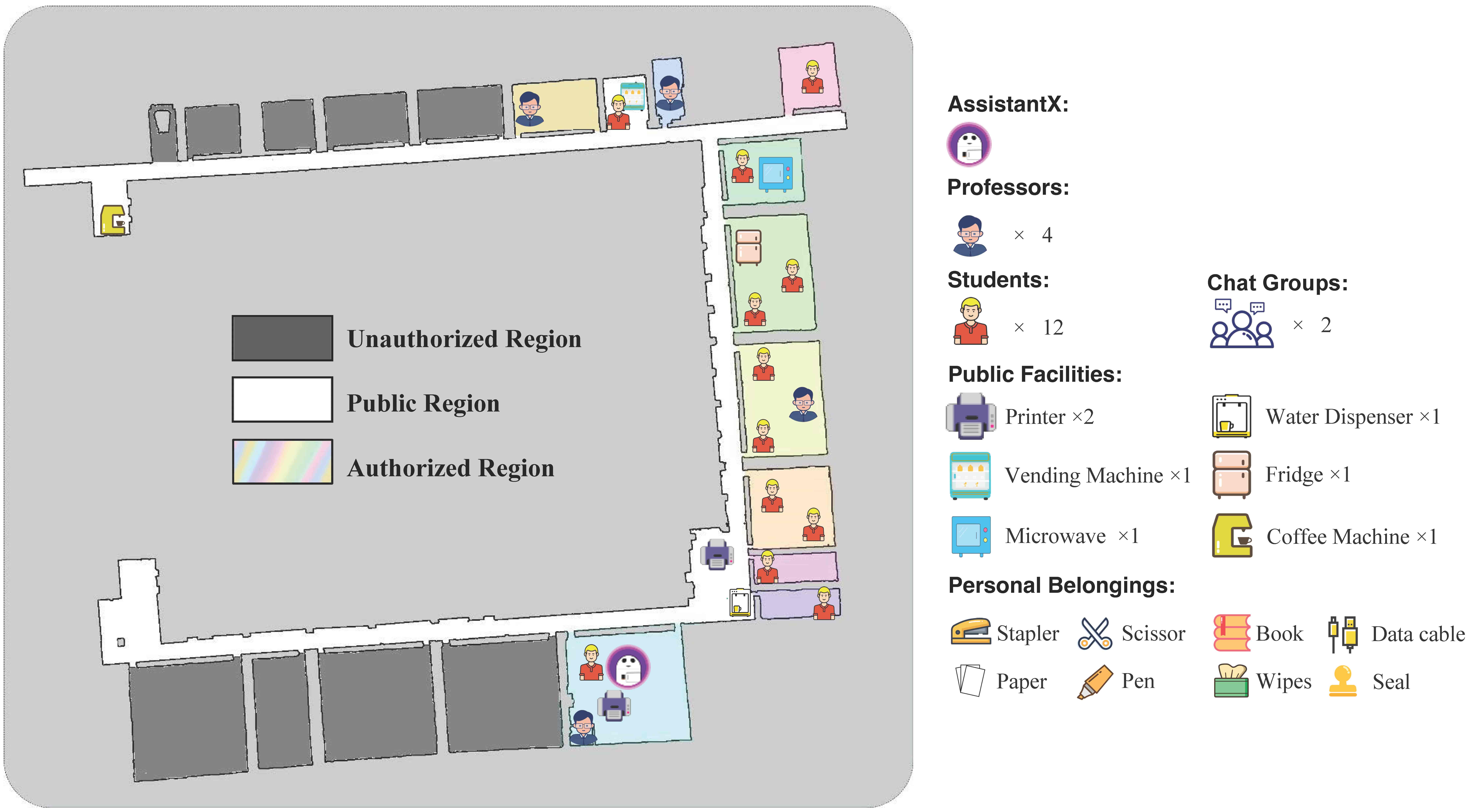}
    \caption{An illustration of our real-world experiment environment, which is also simulated in the simulation experiments.}
    \vspace{-3mm}
    \label{fig:environment}
\end{figure}

\noindent \textbf{Environment.} We built a real-world experimental environment consisting of 23 distinct locations on a semantic map, including 16 workstations and 7 public facilities (see Fig. \ref{fig:environment}). We ensured that every individual possesses at least one personal item, with at least three people sharing the same type of item. For the initial long-term memory set for AssistantX (whose physical details are shown in Fig. \ref{fig:introduction2}), we deliberately included only a subset of ownership information to assess AssistantX's proactive capabilities. Users can issue commands to AssistantX via a one-on-one messaging interface on a social software app. Additionally, we simulated the environment for text-based experiments, where all actions defined in Table \ref{tab:Xpace} are guaranteed to succeed once generated. We also developed a demonstration platform (see Fig. \ref{fig:simulation}) that showcases sample data entries from the dataset, supports online execution, and displays both the inference process and simulations of environmental and robotic state changes. In simulation experiments, an LLM (ChatGPT 4o) reads personnel availability and generates responses for AssistantX, whereas in real-world experiments, human participants interact with AssistantX via their phones based on updated personnel status after each data entry.

\vspace{-2mm}

\begin{table}[h] \large
    \centering
    \caption{The metrics that we used in evaluation.}
    \renewcommand{\arraystretch}{1.2}
    \resizebox{\columnwidth}{!}{%
  
    \begin{tabularx}{\textwidth}{c|c}
        \hline
        \toprule
        \textbf{Evaluation Metric} & \textbf{Description} \\
        \hline
      
        \multicolumn{1}{l|}{\textbf{SR}: Success Rate} & \multicolumn{1}{X}{Success Rate is the percentage of instructions that are successfully completed across various scenarios. }\\
        \hline
      
        \multicolumn{1}{l|}{\textbf{CR}: Completion Rate} & \multicolumn{1}{X}{Calculated by dividing the completed necessary interactions by the total interactions required to complete the task.} \\
        \hline

        \multicolumn{1}{l|}{\textbf{RR}: Redundance Rate} & \multicolumn{1}{X}{Calculated by dividing the number of redundant interactions by the total interactions required.}\\
        
        \hline
         \multicolumn{1}{l|}{\textbf{CTA}: Cyber Task Accuracy} & \multicolumn{1}{X}{The proportion of correct cyber tasks out of the total number of cyber tasks generated.}\\
        
        \hline

        \multicolumn{1}{l|}{\textbf{RTA}: Real-World Task Accuracy} & \multicolumn{1}{X}{The proportion of correct real-world tasks out of the total number of real-world tasks generated.}\\

        \bottomrule
     \hline
     \end{tabularx}%
     }
    \vspace{-3mm}
    \label{tab:eva}
 \end{table}

\begin{figure}[t]
    \centering
    \includegraphics[width=1\linewidth]{image617/white617.jpg}
    \caption{AssistantX is physically embodied as a customized mobile robot equipped with a smart locker, and virtually implemented on a phone with a configured social account.}
    \vspace{-3mm}
    \label{fig:introduction2}
\end{figure}

\begin{table*}[h]
    \centering
    \caption{The results of our method compared to other baselines.}
    \label{tab:compare}
    \renewcommand{\arraystretch}{1.2}
    \setlength{\tabcolsep}{8pt}
    \scalebox{0.8}{
        \begin{threeparttable}
    \begin{tabular}{l c c c c c c c c c c c c c c c}
        \hline
        \toprule
\multicolumn{3}{l}{\hspace{112pt}SR \raisebox{0.1ex}{$\uparrow$}} &
\multicolumn{3}{l}{\hspace{75pt}CR \raisebox{0.1ex}{$\uparrow$}} &
\multicolumn{3}{l}{\hspace{75pt}RR \raisebox{0.1ex}{$\downarrow$}} &
\multicolumn{3}{l}{\hspace{75pt}CTA \raisebox{0.1ex}{$\uparrow$}} &
\multicolumn{3}{l}{\hspace{75pt}RTA
\raisebox{0.1ex}{$\uparrow$}} 
\\
        \cmidrule(lr){2-4}
        \cmidrule(lr){5-7}
        \cmidrule(lr){8-10}
        \cmidrule(lr){11-13}
        \cmidrule(lr){14-16}
        \multicolumn{1}{c}{}
        
        &L1 &L2 &L3 
        &L1 &L2 &L3 
        &L1 &L2 &L3 
        &L1 &L2 &L3 
        &L1 &L2 &L3 \\
        \hline
        
        Direct 
        &0.34 &0.11 &0.06 &0.50
        &0.30 &0.19 &0.09 &0.11
        &0.16 &0.47 &0.28 &0.18
        &0.48 &0.28 &0.19
        \\
        \hline
        CoT \cite{kojima2023largelanguagemodelszeroshot}
        &0.51 &0.27 &0.21 &0.52 &0.43 &0.37 &0.08 &0.11 &0.13 &0.49 &0.40 &0.35 &0.50 &0.41 &0.35
        \\
        \hline
        ReAct \cite{yao2023reactsynergizingreasoningacting}
        &0.57 &0.29 &0.23 &0.56 &0.49 &0.41 &0.04 &0.06 &\textbf{0.07} &0.53 &0.48 &0.39 &0.55 &0.49 &0.40
        \\
        \hline
        Reflexion \cite{shinn2023reflexionlanguageagentsverbal}
        &0.65 &0.52 &0.42 &0.74 &0.59 &0.58 &0.05 &0.08 &0.10 &0.73 &0.57 &0.47 &0.67 &0.55 &0.44
        \\
        \hline
        Mobile-Agent-v2 \cite{wang2024mobileagentv2mobiledeviceoperation}
        &0.70 &0.62 &0.52 &0.76 &0.63 &0.54 &0.06 &0.07 &0.08 &0.75 &0.60 &0.51 &0.72 &0.60 &0.49
        \\
        \hline
        Ours 
        &\textbf{0.81} &\textbf{0.74} &\textbf{0.66} &\textbf{0.83} &\textbf{0.73} &\textbf{0.62} &\textbf{0.03} &\textbf{0.05} &0.08 &\textbf{0.85} &\textbf{0.70} &\textbf{0.61} &\textbf{0.82} &\textbf{0.71} &\textbf{0.60}
        \\
        \bottomrule
        \hline
    \end{tabular}
     \begin{tablenotes}    
        \footnotesize              
        \item[*] The bold values indicate that the method achieves the best performance for a specific metric at a certain difficulty level.
        \end{tablenotes}            
        \end{threeparttable}
    }
    \vspace{-3mm}
\end{table*}

\begin{table*}[h]
    \centering
    \caption{The results of ablation study.}
    \label{tab:ablation}
    \renewcommand{\arraystretch}{1.2}
    \setlength{\tabcolsep}{8pt}
    \scalebox{0.9}{
        \begin{threeparttable}
    \begin{tabular}{l|c c c c c c c c c c c c }
        \hline
        \toprule
        \multicolumn{1}{l}{\multirow{1}{*}{\textbf{Perception Agent}}}
            &\multicolumn{3}{c}{\color{darkgreen}\CheckmarkBold}
            &\multicolumn{3}{c}{\color{darkred}\XSolidBrush}
            &\multicolumn{3}{c}{\color{darkgreen}\CheckmarkBold}
            &\multicolumn{3}{c}{\color{darkgreen}\CheckmarkBold}
            \\
            \multicolumn{1}{l}{\multirow{1}{*}{\textbf{Reflection Agent}}}
            &\multicolumn{3}{c}{\color{darkgreen}\CheckmarkBold}
            &\multicolumn{3}{c}{\color{darkgreen}\CheckmarkBold}
            &\multicolumn{3}{c}{\color{darkred}\XSolidBrush}
            &\multicolumn{3}{c}{\color{darkgreen}\CheckmarkBold}
            \\
            \multicolumn{1}{l}{\multirow{1}{*}{\textbf{Planning Agent}}}
            &\multicolumn{3}{c}{\color{darkgreen}\CheckmarkBold}
            &\multicolumn{3}{c}{\color{darkgreen}\CheckmarkBold}
            &\multicolumn{3}{c}{\color{darkgreen}\CheckmarkBold}
            &\multicolumn{3}{c}{\color{darkred}\XSolidBrush}
            \\
        \cmidrule(lr){2-4}
        \cmidrule(lr){5-7}
        \cmidrule(lr){8-10}
        \cmidrule(lr){11-13}
        \multicolumn{1}{c}{}
        
        &L1 &L2 &L3 
        &L1 &L2 &L3 
        &L1 &L2 &L3 
        &L1 &L2 &L3 \\
        \hline
        
        SR {$\uparrow$}
        &\textbf{0.81} &\textbf{0.74} &\textbf{0.66} &0.80
        &0.57 &0.34 &0.70 &0.48
        &0.45 &0.64 &0.44 &0.16
        \\
        \hline
        CR {$\uparrow$}
        &\textbf{0.83} &\textbf{0.73} &\textbf{0.62} &0.81
        &0.65 &0.49 &0.78 &0.63
        &0.57 &0.71 &0.53 &0.25
        \\
        \hline
        RR {$\downarrow$}
        &\textbf{0.03} &0.05 &\textbf{0.08} &0.05 &0.08
        &0.10 &\textbf{0.03}
        &\textbf{0.04} &0.09 &0.04 &0.09 &0.12
        \\
        \hline
        CTA {$\uparrow$}
        &\textbf{0.85} &\textbf{0.70} &\textbf{0.61} &0.80
        &0.63 &0.47 &0.77 &0.62
        &0.56 &0.71 &0.53 &0.25\\
        \hline
        RTA {$\uparrow$}
        &\textbf{0.82} &\textbf{0.71} &\textbf{0.60} &0.80
        &0.64 &0.48 &0.77 &0.63
        &0.57 &0.72 &0.52 &0.25
        \\
        \bottomrule
        \hline
    \end{tabular}
     \begin{tablenotes}    
        \footnotesize              
        \item[*]   The bold values indicate that the method achieves the best performance for a specific metric at a certain difficulty level.
        \end{tablenotes}            
        \end{threeparttable}
    }
    \vspace{-5mm}
\end{table*}

\begin{table}[h] \small
    \centering
    \caption{The evaluation of LLM backbones.}
    \begin{tabular}{cccc}
        \toprule
        Base model & Gemini-1.5-pro & Claude-3.5-Sonnet & GPT-4o \\
        \hline
        SR & 0.70 & 0.68 & 0.72\\
        \hline
        CR & 0.73 & 0.70 & 0.76\\
        \hline
        RR & 0.07 & 0.06 & 0.05\\
        \bottomrule
    \end{tabular}
    \vspace{-1mm}
    \label{tab:backbone}
 \end{table}

\begin{figure}[t]
    \centering
    \includegraphics[width=1.0\linewidth]{image617/web617.jpg}
    \caption{The platform used for simulation and display.}
    \vspace{-7mm}
    \label{fig:simulation}
\end{figure} 

\noindent \textbf{Dataset.} Based on survey responses from over 300 students and faculty members regarding daily tasks and errands they found exhausting and wished to automate, we developed a dataset tailored to our environmental setting to rigorously evaluate the effectiveness of our approach. The dataset comprises instruction content and personnel status information, including 30 base instructions (with all individuals marked as available) and 180 variants where the instruction content remains identical to base instructions but one or more personnel are marked unavailable. Each entry includes a feature indicating whether the instruction is achievable, which the robot must identify during execution. For instance, if person A is unavailable to sign a file or if no one with a pen is available, the instruction is deemed unachievable. The dataset is categorized into three difficulty levels: L1 tasks can be completed reactively based solely on user instructions; L2 tasks involve unavailable personnel, requiring the robot to autonomously search for alternatives; and L3 tasks necessitate engaging with other humans for assistance when all relevant personnel in long-term memory are unavailable. Comprehensive details of the dataset are presented in TABLE~\ref{tab:dataset}, with specific examples shown on our simulation platform (see Fig. \ref{fig:simulation}).

\vspace{1mm}

\noindent \textbf{Baselines.} To rigorously evaluate our method, we compared it with 5 LLM-powered agentic baselines, ensuring that they were based on the same model and that the prompts were all tailored to our environment setting to maintain fairness: 
\begin{itemize}
    \item \textbf{Direct}: Directly generating tasks based on user instructions, with an example provided for in-context learning.
    \item \textbf{CoT} \cite{kojima2023largelanguagemodelszeroshot}: Build on Direct by encouraging step-by-step reasoning, enhancing structured thinking.
    \item \textbf{ReAct} \cite{yao2023reactsynergizingreasoningacting}: Integrating explicit environmental observation and thinking stages into reasoning, generating tasks iteratively through cycles of thought, action, and observation.
    \item \textbf{Reflexion} \cite{shinn2023reflexionlanguageagentsverbal}: Similar to ReAct but includes reflection on the results of actions, improving reasoning through high-level explanations of the executed tasks outcomes.
    \item \textbf{Mobile-Agent-v2} \cite{wang2024mobileagentv2mobiledeviceoperation}: Involving three agents and one perception module, including a reflection agent, but struggles with long sequential tasks due to its full-scale perception mechanisms, which result in low efficiency and excessive irrelevant information. 
\end{itemize}

\begin{figure*}[t]
    \centering
    \includegraphics[width=1.0\linewidth]{image617/CaseStudy.jpg}
    \caption{We demonstrate that AssistantX can reactively respond to Lee’s request and operate autonomously. When Mao is unavailable for printing, it actively searches memory for alternatives, identifying Wu. When Wu is also unavailable, AssistantX proactively seeks help in an active group chat to complete the complex task with human collaboration. Two representative inference processes showcasing the generation of proactive thoughts and behaviors are also presented.}
    \vspace{-4mm}
    \label{fig:casestudy}
\end{figure*}

\vspace{-1mm}

\subsection{Evaluation}\label{subsec:evaluation}

To assess the effectiveness of our framework, we define 5 evaluation metrics in TABLE \ref{tab:eva}. We choose ChatGPT-4o as the foundation model, with detailed model baseline comparison results provided in TABLE \ref{tab:backbone}. Our dataset of 210 entries was run 5 times in simulation, and the average of the 5 runs was computed for each metric.

The overall evaluation presented in Table \ref{tab:compare} highlights the robustness of our approach relative to all baselines, particularly in complex tasks, and demonstrates its ability to function effectively in dynamic environments. In terms of SR, our method achieves 0.81, 0.74, and 0.66 for L1, L2, and L3, respectively—exceeding Mobile-Agent-v2 (the second-best) by 0.11, 0.12, and 0.14. Moreover, CR is notably higher, reaching 0.83 at L1 compared to 0.76 for Mobile-Agent-v2, a 0.07 improvement even in the simplest tasks. Our method also exhibits the lower RR, ensuring higher efficiency in task execution. In terms of CTA and RTA, our approach leads with improvements of up to 0.10 at least. The overall results further indicate that single-agent methods, such as Direct and CoT, are less effective than multi-agent approaches like Mobile-Agent-v2 and Reflexion, even with step-by-step prompting. However, we also observed that incorporating the reflection mechanism increases the redundancy rate, with the non-reflective ReAct achieving the lowest L3 redundancy rate of 0.07. This is probably attributable to the additional reasoning steps or errors introduced during reflection processes.

To further validate the effectiveness of each agent, we conducted ablation experiments (see Table \ref{tab:ablation}). Our findings indicate that when Perception Agent and Planning Agent are individually removed, performance degrades notably, especially in complex tasks (L3). Without Perception Agent, SR and CR decrease slightly at L2 but drop significantly at L3 (SR: 0.66 to 0.34, CR: 0.62 to 0.49). Removing the Planning Agent causes a substantial decline at all difficulty levels: SR drops to 0.64/0.44/0.16 and CR to 0.71/0.53/0.25 at L1/L2/L3, while RR increases to 0.04/0.09/0.12. The removal of Reflection Agent leads to milder degradation in SR and CR, but also results in a lower RR at L1 and L2, suggesting that while Reflection Agent improves accuracy, the added reasoning steps may sometimes induce redundant actions.

\subsection{Real-World Experiment}\label{sec:real-world} 

We further evaluated AssistantX in our real lab to assess its effectiveness in streamlining workflows and enhancing productivity. We present a complete scenario (see Fig.~\ref{fig:casestudy}), highlighting its capabilities in: (1) \textit{reactively} responding to user instructions, autonomously delivering items, and notifying relevant personnel online; (2) \textit{actively} locating an available individual with a pen when the personnel lack one for signing; and (3) \textit{proactively} seeking assistance in group chats when no personnel are available for printing. 

AssistantX was also deployed for open use by researchers not affiliated with the development team for one and a half months. To evaluate its real-world impact, we conducted a user study aimed at assessing the system’s ability to reduce physical burden, improve task efficiency, and support seamless human-robot collaboration in an unsupervised setting. After each interaction, a short survey was delivered via AssistantX’s integrated messaging system, consisting of multiple-choice questions (with \textit{Yes}, \textit{No}, and \textit{Unsure} options) covering task success, walking reduction, user satisfaction, and perceived productivity, along with an optional open-ended field for free-form feedback. In total, we collected 289 valid responses.

\begin{figure}[t]
    \centering
    \includegraphics[width=1.0\linewidth]{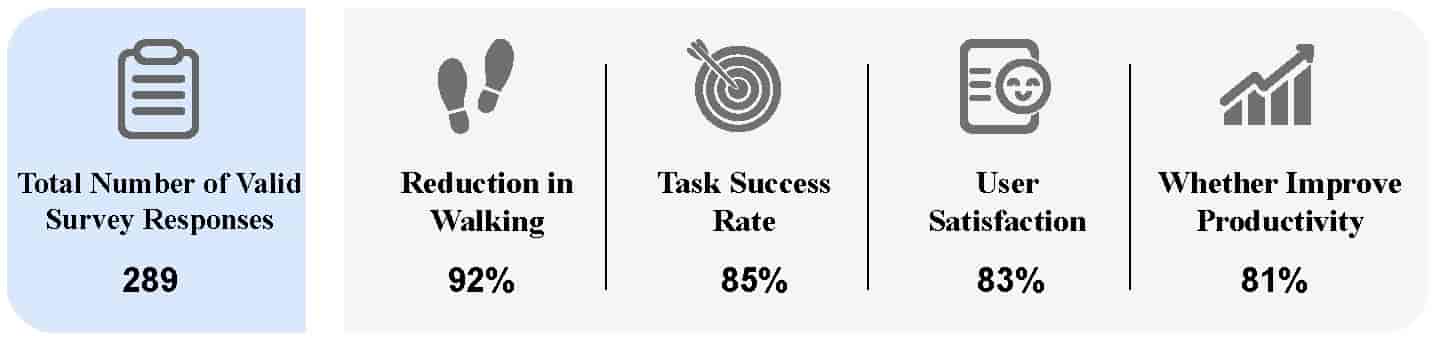}
    \caption{Aggregated results from 289 valid user survey responses collected during the real-world deployment of AssistantX, indicating strong acceptance and measurable benefits across physical, functional, and cognitive dimensions.}
    \vspace{-7mm}
    \label{fig:survey}
\end{figure}

The aggregated results, shown in Fig.~\ref{fig:survey}, indicate that 92\% of users experienced a reduction in walking effort, 85\% reported successful task completion, 83\% expressed satisfaction with AssistantX, and 81\% affirmed that it improved their productivity. These outcomes suggest the system was able to effectively integrate into natural workflows, offering measurable value in task support and reducing user workload. The system's consistent performance across a range of users and tasks points to its robustness and adaptability in real-world human environments.

Beyond the quantitative metrics, the deployment of AssistantX revealed several qualitative benefits that underscore the value of autonomous embodied assistants in everyday collaborative environments. By serving as an always-available intermediary for routine yet interruptive tasks—such as delivering items, locating personnel, or sending online notifications—AssistantX reduced cognitive load and minimized workflow fragmentation. Users reported that the robot's presence enabled them to stay focused on primary tasks without frequent context-switching, effectively improving concentration and task continuity. In one notable case, a user suffering from plantar fasciitis highlighted the robot's assistance as significantly beneficial in reducing unnecessary movement. Such feedback demonstrates the potential impact of robots like AssistantX in accessibility-related use cases, including support for elderly or mobility-impaired individuals. Given its lightweight design and generalizable task structure, our method is well-suited for rapid deployment to other robotic platforms and assistive contexts.

Moreover, the system’s ability to strategically engage humans only when needed, rather than attempting full autonomy, fostered a sense of cooperative fluency rather than resistance. This selective involvement aligns with the notion of ``precision-activated collaboration,'' where human assistance is invoked at critical points rather than ubiquitously. Informal user feedback also suggested that the system encouraged a more equitable distribution of responsibilities in shared spaces by offloading low-complexity but high-frequency tasks from individuals to the robot. These observations support the emerging view that embodied agents, when properly scoped and socially integrated, can enhance not only operational efficiency but also the overall social dynamics and inclusivity of collaborative work environments.

\section{CONCLUSION}\label{sec:conclusion}

In this study, we present AssistantX, an LLM-powered proactive assistant, designed to operate autonomously in a real-world office environment. By leveraging the PPDR4X framework, we endowed AssistantX with the ability to autonomously interpret, plan, and execute both cyber and real-world actions, significantly enhancing operational efficiency. The experimental results substantiate the feasibility of our framework, opening up new avenues for its application across various domains. Moreover, our approach demonstrates remarkable robustness and scalability, exploring new paradigms of human-robot collaboration. Future work will focus on refining AssistantX’s natural language understanding capabilities, expanding its repertoire of physical interactions, and exploring its scalability within more intricate and expansive environments. Our work underscores key technological and conceptual pathways for advancing next-generation autonomous assistants to enhance operational efficiency, cognitive support, and productivity, ultimately aiming to develop intelligent, adaptive systems that integrate seamlessly into daily workflows and redefine human-agent interaction across both virtual and real-world environments.

\addtolength{\textheight}{-10cm}  



\bibliographystyle{IEEEtran} 
\bibliography{reference}
\end{document}